\documentclass[conference]{IEEEtran}
\IEEEoverridecommandlockouts
\usepackage{cite}
\usepackage{amsmath,amssymb,amsfonts}
\usepackage{algorithmic}
\usepackage{graphicx}
\usepackage{textcomp}
\usepackage{xcolor}
\def\BibTeX{{\rm B\kern-.05em{\sc i\kern-.025em b}\kern-.08em
    T\kern-.1667em\lower.7ex\hbox{E}\kern-.125emX}}
\usepackage{orcidlink}
\usepackage{marvosym}
\usepackage{geometry}
\usepackage{array}
\usepackage{threeparttable}
\usepackage{multirow}
\usepackage{amsmath}
\usepackage{pifont}   
\newcommand{\cmark}{\ding{51}}
\newcommand{\xmark}{\ding{55}}
\usepackage{hyperref}
\usepackage{booktabs}
\usepackage{makecell}

\begin{document}

\title{

MG-Grasp: Metric-Scale Geometric 6-DoF Grasping Framework with Sparse RGB Observations
}

\author{
Kangxu Wang$^{1}$, 
Siang Chen$^{1}$, 
Chenxing Jiang$^{2}$, 
Shaojie Shen$^{2}$, 
Yixiang Dai$^{1,\dag}$,
Guijin Wang$^{1}$
\thanks{$^{1}$The Department of Electronic Engineering, Tsinghua University, Beijing 100084, China.}
\thanks{$^{2}$The Department of Electronic and Computer Engineering, The Hong Kong University of Science and Technology, Hong Kong, China.}
\thanks{$^{\dagger}$Corresponding Author: {\tt daiyx23@mails.tsinghua.edu.cn.}}
\vspace{-15pt}
}

\maketitle

\begin{abstract}

Single-view RGB-D grasp detection remains a common choice in 6-DoF robotic grasping systems, which typically requires a depth sensor. While RGB-only 6-DoF grasp methods has been studied recently, their inaccurate geometric representation is not directly suitable for physically reliable robotic manipulation, thereby hindering reliable grasp generation. To address these limitations, we propose MG-Grasp, a novel depth-free 6-DoF grasping framework that achieves high-quality object grasping. Leveraging two-view 3D foundation model with camera intrinsic/extrinsic, our method reconstructs metric-scale and multi-view consistent dense point clouds from sparse RGB images and generates stable 6-DoF grasp. Experiments on GraspNet-1Billion dataset and real world demonstrate that MG-Grasp achieves state-of-the-art (SOTA) grasp performance among RGB-based 6-DoF grasping methods. 

\end{abstract}


\section{INTRODUCTION}
Robotic 6-DoF grasping is a core capability in fields such as intelligent manufacturing, domestic service, and medical treatment. In early grasp pose detection pipelines, the task was commonly formulated as predicting full 6-DoF gripper poses directly from 3D observations~\cite{ten2017grasp}. Recent learning-based 6-DoF grasp detectors have achieved strong performance, and large-scale benchmarks such as GraspNet-1Billion~\cite{fang2020graspnet} have further accelerated progress and standardized evaluation. Representative advances on this benchmark continue to improve grasp quality and efficiency, including graspness-guided methods~\cite{wang2021graspness}, heatmap-guided detection~\cite{chen2023efficient}, and flexible local grasp modeling~\cite{xie2026rethinking}. However, most existing high-performing 6-DoF pipelines still assume access to explicit 3D geometry that is typically obtained via depth sensing, as in representative RGB-D approaches~\cite{sundermeyer2021contact,wang2021graspness, chen2023efficient, xie2026rethinking, breyer2021volumetric,fang2020graspnet}, which can increase system cost and complicate large-scale deployment in real factories. This motivates RGB-only (depth-free) 6-DoF grasping, aiming to generate physically reliable grasps from ubiquitous multi-view RGB observations without depth hardware.

Recent progress in neural rendering ~\cite{mildenhall2021nerf,chen2021mvsnerf,liu2022neural,muller2022instant,deng2020jaxnerf} has inspired a line of RGB-only grasping methods that reconstruct geometry from multi-view RGB and then synthesize 6-DoF grasps, as GraspNeRF~\cite{dai2023graspnerf}, RGBGrasp\cite{liu2024rgbgrasp} and NeuGrasp~\cite{fan2025neugrasp}. Despite promising results, their geometry quality under sparse views can be insufficient for capturing fine-grained contact details, thereby limiting physically reliable grasp generation. In parallel, feed-forward 3D foundation models have rapidly advanced RGB-only geometry prediction, ranging from two-view prediction in DUSt3R~\cite{wang2024dust3r} and MASt3R~\cite{leroy2024grounding} to more general multi-view geometry prediction in VGGT~\cite{wang2025vggt} and related architectures \cite{keetha2025mapanything}. Building on these models, recent grasping systems such as GraspView\cite{wang2025graspview} and SparseGrasp\cite{yu2024sparsegrasp} demonstrate the feasibility of coupling foundation geometry model with downstream grasp execution. However, in 6-DoF grasping scenarios with known camera parameters, their predicted geometry can still be non-metric and cross-view inconsistent with scale alignment. An alternative direction is Multi-View Stereo (MVS), where classical pipelines \cite{schonberger2016structure} and learning-based backbones~\cite{yao2018mvsnet,Gu_2020_CVPR, izquierdo2025mvsanywhere} reconstruct dense 3D structure from posed RGB images. Such geometry has been adopted in 6-DoF grasping systems\cite{lin2021multi, avigal20206, saikia2025robotic}. Nevertheless, MVS-based methods require sufficient visual overlap and predefined depth ranges, and their effectiveness in tabletop scenarios with multiple viewpoints is not satisfactory.



\begin{figure*}[h]
  \centering
  \includegraphics[width=1.0\linewidth]{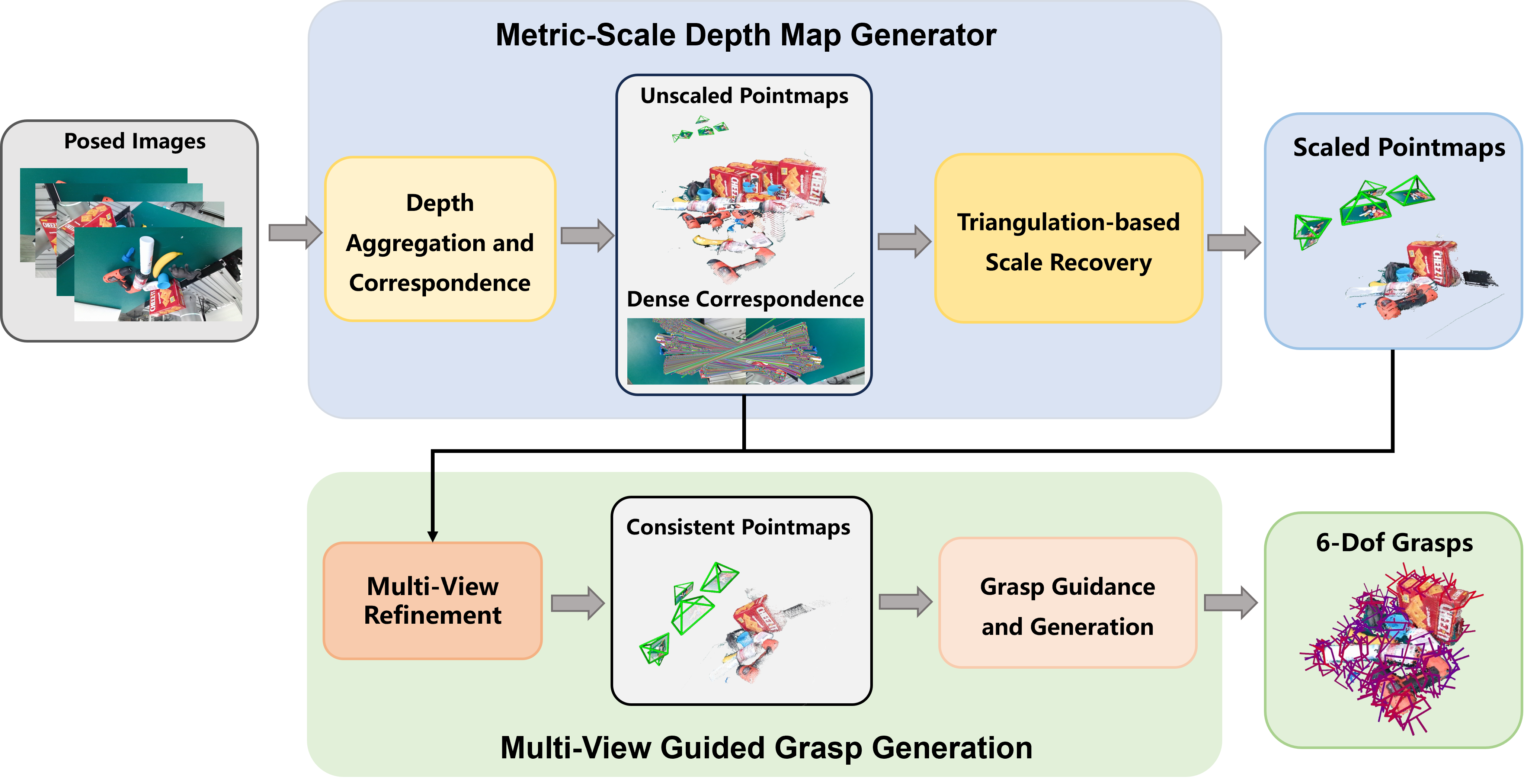}
  \caption{Pipeline of MG-Grasp:  given sparse posed RGB images, we first perform Depth Aggregation and Dense Correspondence to obtain up-to-scale pointmaps, and then recover metric scale via Triangulation-based Scale Recovery. The metric pointmaps are further refined with multi-view consistency optimization to produce consistent geometry, which is finally used for grasp guidance and 6-DoF grasp generation.
}
  \label{pipeline_grasp}
\end{figure*}

To address these challenges, we propose \textbf{MG-Grasp}, a \textbf{M}etric-Scale \textbf{G}eometric 6-DoF \textbf{Grasp}ing Framework with Sparse RGB Observations. Rather than treating multi-view RGB reconstruction as a coarse pre-processing step, MG-Grasp designs metric geometry recovery, multi-view consistency refinement, and couples grasp generation into a unified pipeline tailored for robotic 6-DoF grasping. Given sparse multi-view RGB images with known camera intrinsics and extrinsics, we first convert two-view, up-to-scale depth estimates into metric depth by triangulation-based scale grounding, and refine all views through a confidence-weighted two-stage multi-view optimization. Finally, we fuse a grasp-oriented, reliability-filtered point cloud and decode stable 6-DoF grasps with grasp model. Our pipeline achieves a seconds-level end-to-end latency, making it practical for deployment under sparse-view settings. Extensive experiments on GraspNet and real-world setups demonstrate strong performance among RGB-only 6-DoF grasping methods, with notable robustness under sparse-view inputs. Our main contributions can be summarized as follows:
\begin{itemize}
\item We propose a depth-free 6-DoF grasping framework, achieving high-quality, physically reliable grasping with sparse multi-view observations.
\item We develop a Triangulation-based Scale Recovery scheme that grounds up-to-scale two-view predictions into a global metric coordinate. 
\item We design a two-stage confidence-weighted Multi-View Refinement module that enforces dense cross-view consistency, yielding grasp-oriented, multi-view consistent geometry for reliable grasp generation.
\end{itemize}





\section{Related Work}

\subsection{Scene-Level Grasping from RGB-D}

Early 6-DoF grasp detection methods commonly follow a sampling-and-evaluation pipeline~\cite{ten2017grasp,liang2019pointnetgpd}, where dense grasp candidates are generated and scored by point-cloud classifiers, albeit with limited efficiency. GraspNet~\cite{fang2020graspnet} improves scalability by predicting grasps for all points in a scene in a more unified manner, and subsequent works further enhance grasp quality via region-based refinement~\cite{zhao2021regnet}, transformer-based point modeling~\cite{liu2022transgrasp}, graspness-driven proposal learning~\cite{wang2021graspness}, and heatmap-guided generation~\cite{chen2023efficient}. More recently, FlexLoG~\cite{xie2026rethinking} adopts a grasp-centric formulation that decouples guidance from generation, achieving strong generalization to unseen objects. 

However, all these methods rely on high-quality depth sensors, which increase system cost and complicate large-scale deployment in real environments. In contrast, our method leverages only sparse multi-view RGB images for 3D reconstruction, eliminating the need for depth cameras while maintaining geometric accuracy through explicit multi-view constraints.

\subsection{RGB-only 6-DoF Grasping}

Recent studies explore RGB-only 6-DoF grasping by leveraging multi-view RGB observations to recover grasp-relevant 3D cues without depth sensors. NeRF-based pipelines couple implicit multi-view reconstruction with grasp prediction, including GraspNeRF~\cite{dai2023graspnerf}, RGBGrasp~\cite{liu2024rgbgrasp}, and NeuGrasp~\cite{fan2025neugrasp}. In parallel, GraspView~\cite{wang2025graspview} studies RGB-only grasping with active perception, using reconstructed geometry predicted by VGGT\cite{wang2025vggt} and aligned to metric scale with robot kinematics. SparseGrasp\cite{yu2024sparsegrasp} and VG-Grasp\cite{dai2026visual} utilizes DUSt3R\cite{wang2024dust3r} to generate point cloud for grasp generation. Lin's work\cite{lin2021multi}  reconstructs dense point clouds with CasMVSNet\cite{Gu_2020_CVPR} for robotic manipulation, and Avigal's work\cite{avigal20206} use a learnt stereo machine to predict depth maps from multiple RGB views and select grasps via a multi-view GQ-CNN policy.

Compared with these RGB-only pipelines, MG-Grasp does not treat multi-view RGB reconstruction as a coarse pre-processing step. Leveraging known camera parameters, our method couples metric scale recovery, cross-view consistency optimization and grasp generation into a unified pipeline, yielding physically reliable 6-DoF grasps.

\section{Method}

\subsection{Overview}

Our goal is to recover metric-scale and cross-view consistent geometry from sparse posed RGB views, and directly steer the geometry toward reliable 6-DoF grasp generation. As shown in Fig. \ref{pipeline_grasp}, given multi-view posed RGB images as input, our framework first processes them through \textbf{Metric-Scale Depth Map Generator} module. We recover physical-world scale to produce metric depth maps through a two-view geometry module. These metric depth maps, along with the original images, are then fed into the \textbf{Multi-View Guided Grasp Generation} module. We perform dense multi-view optimization to refine the initial depth estimates into multi-view consistent depth maps. The refined depth maps are then unprojected, sampled and implicitly decoded to 6-DoF grasp poses via an Local Grasp Model. This pipeline ensures that the geometric refinements directly contribute to producing successful grasps, resulting in a high-precision system that outputs both a consistent 3D structure and actionable grasp configurations from sparse multi-view RGB images.

\subsection{Metric-Scale Depth Map Generator}
\label{subsec:metric_generator}
This module generates metric-scale depth maps from multi-view RGB images, providing a strong initialization for the subsequent module. We first obtain per-view depth estimates from a two-view geometry model, which are accurate but inherently up-to-scale. Leveraging known camera poses, we then recover metric scale by triangulating matched pixels across views and aligning the predicted depths to the triangulated depths. As a result, the depth initialization is grounded in a common physical coordinate system.

\textbf{Depth Aggregation and Correspondence.} 
We leverage MASt3R~\cite{leroy2024grounding} as a two-view geometry model. For each ordered pair $(i,j)$ and images \(\mathbf{I}_i, \mathbf{I}_j \in \mathbb{R}^{H \times W \times 3}\) , we obtain
\begin{align}
(\mathbf{X}_i^i, \mathbf{X}_i^j, \mathbf{C}_i^i, \mathbf{C}_i^j, \mathbf{D}_i^i, \mathbf{D}_i^j, \mathbf{Q}_i^i, \mathbf{Q}_i^j)
= \text{MASt3R}(\mathbf{I}_i, \mathbf{I}_j),
\end{align}
where \(\mathbf{X}\in \mathbb{R}^{H \times W \times 3}\) denotes pointmaps with their confidences \(\mathbf{C} \in \mathbb{R}^{H \times W \times 1}\), \(\mathbf{D} \in \mathbb{R}^{H \times W \times d}\) denotes feature maps, and corresponding feature confidences \(\mathbf{Q} \in \mathbb{R}^{H \times W \times 1}\). $\mathbf{X}_i^j$ denotes the pointmap of image $j$ represented in the coordinate frame of camera $i$. The same convention applies to $\mathbf{C}$, $\mathbf{D}$, and $\mathbf{Q}$.

For each view \(i\), we compute confidence-weighted aggregation of depth predictions across all pairs using \(\mathbf{C}_i^i\) as weights, yielding an initial depth map \(\hat{\mathbf{Z}}_i\). Meanwhile, we average \(\mathbf{C}_i^i\) across all pairs to obtain an initial confidence map \(\hat{\mathbf{C}}_i\), which reflects the average reliability of each pixel over all observations.

To establish dense correspondences across views and ensure matching speed and density, following MASt3R-SLAM~\cite{murai2025mast3r}, we compute per-pixel dense matching
\begin{align}
\mathbf{M}_{i\rightarrow j} = \mathcal{M}\big(\mathbf{X}_i^i, \mathbf{X}_i^j, \mathbf{D}_i^i, \mathbf{D}_i^j\big),
\end{align}
between views $(i,j)$. Due to partial overlap and ambiguous matches, we retain only reliable correspondences. For a candidate correspondence $(\mathbf{x}_i^k, \mathbf{x}_j^k)\in \mathbf{M}_{i\rightarrow j}$, matching confidence is defined as
\begin{align}
\mathbf{Q}_{i\rightarrow j}(\mathbf{x}_i^k, \mathbf{x}_j^k)
= \sqrt{\mathbf{Q}_i^i(\mathbf{x}_i^k)\, \mathbf{Q}_i^j(\mathbf{x}_j^k)}.
\end{align}
We then filter correspondences with a threshold $\tau_Q$ and obtain the valid set:
\begin{align}
\mathbf{M}_{i\rightarrow j}^v = \left\{
(\mathbf{x}_i^k,\mathbf{x}_j^k)\in \mathbf{M}_{i\rightarrow j}
\ \big|\ 
\mathbf{Q}_{i\rightarrow j}(\mathbf{x}_i^k,\mathbf{x}_j^k)>\tau_Q
\right\},
\end{align}
and denote by $\mathbf{Q}_{i\rightarrow j}^v$ the confidence score of each retained correspondence.

\textbf{Triangulation-based Scale Recovery.}
As shown in Fig.~\ref{pipeline_grasp}, the initial depth map \(\hat{\mathbf{Z}}_i\) suffers from a severe scale ambiguity and thus lacks metric meaning. To recover metric scale across views, we ground multi-view depths in a common physical coordinate system via triangulation. Different from \cite{wang2024dust3r, duisterhof2025mast3r} that typically incorporates the per-view scale into a joint optimization together with other variables, we decouple metric scale recovery from the subsequent refinement process, which in turn benefits the downstream module. 

We select a minimal set of view pairs that covers all views and has sufficient overlap, by choosing for each view its best-matched neighbor according to correspondence confidence. For each valid correspondence $(\mathbf{x}_i^k,\mathbf{x}_j^k)\in \mathbf{M}_{i\rightarrow j}^v$, we triangulate the corresponding 3D point in the world frame using known poses $\mathbf{T}_i,\mathbf{T}_j$:
\begin{align}
\mathbf{p}_w^k = \text{Tri}(\mathbf{x}_i^k, \mathbf{x}_j^k, \mathbf{K}, \mathbf{T}_i, \mathbf{T}_j),
\end{align}
where \(\text{Tri}(\cdot)\) denotes the triangulation operator and \(\mathbf{p}_w^k\) is the resulting 3D point in the world frame. We then transform \(\mathbf{p}_w^k\) into each camera frame to obtain metric depth estimates \(\mathbf{Z}_{i}^{\text{tri}}(\mathbf{x}_i^k)\) and \(\mathbf{Z}_{j}^{\text{tri}}(\mathbf{x}_j^k)\).

We recover the per-view scale factors by aligning the predicted depths to the triangulated metric depths over valid correspondences:
\begin{equation}
\begin{aligned}
s_i = 
\frac{\sum_{(\mathbf{x}_i^k,\mathbf{x}_j^k)\in \mathbf{M}_{i\rightarrow j}^v} \mathbf{Z}_{i}^{\text{tri}}(\mathbf{x}_i^k)}
{\sum_{(\mathbf{x}_i^k,\mathbf{x}_j^k)\in \mathbf{M}_{i\rightarrow j}^v} \hat{\mathbf{Z}}_{i}(\mathbf{x}_i^k)}, \\
s_j = 
\frac{\sum_{(\mathbf{x}_i^k,\mathbf{x}_j^k)\in \mathbf{M}_{i\rightarrow j}^v} \mathbf{Z}_{j}^{\text{tri}}(\mathbf{x}_j^k)}
{\sum_{(\mathbf{x}_i^k,\mathbf{x}_j^k)\in \mathbf{M}_{i\rightarrow j}^v} \hat{\mathbf{Z}}_{j}(\mathbf{x}_j^k)}.
\end{aligned}
\end{equation}
Finally, the metric-scale depth maps are obtained as:
\begin{align}
\tilde{\mathbf{Z}}_i = s_i \hat{\mathbf{Z}}_i,
\quad
\tilde{\mathbf{Z}}_j = s_j \hat{\mathbf{Z}}_j.
\end{align}

\subsection{Multi-View Guided Grasp Generation}  
\label{subsec:refine_and_grasp}
\begin{figure}[h]
  \centering
  \includegraphics[width=0.5\textwidth]{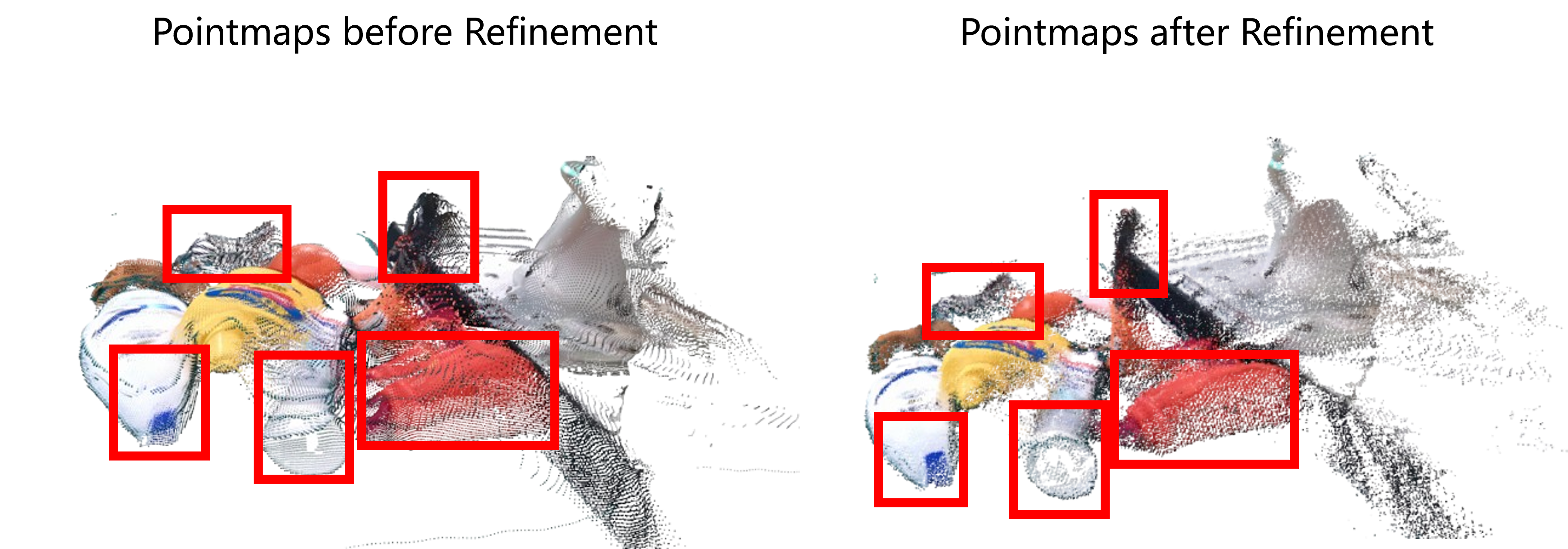}
  \caption{Left: metrically scaled pointmaps before refinement; right: refined pointmaps. The highlighted regions (red boxes) illustrate misaligned surface layers before refinement, which are largely removed after refinement, resulting in multi-view consistent geometry for grasping.}
  \label{refinement_diff}
\end{figure}
This module refines metric depth to be multi-view consistent and decodes stable 6-DoF grasps from the refined geometry. We adopt an optimization-based refinement scheme driven by dense correspondences and their confidence scores. Leveraging the refined depth maps, we unproject them into a multi-view consistent point cloud, which is then sampled and implicitly decoded into 6-DoF grasp via a Local Grasp Model.

\textbf{Multi-View Refinement.} As shown in Fig. \ref{refinement_diff}, the directly unprojected metric-scale depth maps $\tilde{\mathbf{Z}}$ exhibit severe misalignment across views. Given camera intrinsics \(\mathbf{K}\) and poses \(\mathbf{T}\in \mathrm{SE(3)}\), metrically scaled initialization \(\mathbf{\tilde{Z}} = \{\mathbf{\tilde{Z}}_i\}_{i=0}^{N-1}\), the valid dense correspondences \(\mathbf{M}^v\), and their associated confidence scores \(\mathbf{Q}^v\), we employ robust loss to mitigate residual outliers. We optimize \(\mathbf{Z} = \{\mathbf{Z}_i\}_{i=0}^{N-1}\) initialized from $\mathbf{\tilde{Z}}$. Inspired by the sequential optimization paradigm in LIVO systems\cite{zheng2024fast, zhou2025lir}, where a 3D-consistency objective provides a stable initialization before enforcing tighter reprojection consistency, we adopt a two-stage confidence-weighted refinement strategy.

\paragraph{Stage I: 3D consistency}
For each correspondence \((\mathbf{x}_i^k,\mathbf{x}_j^k)\in \mathbf{M}_{i\rightarrow j}^v\) in image pair \((i,j)\), we back-project them into the world frame as:
\begin{equation}
\begin{aligned}
\mathbf{p}_{w,i}^k = \pi^{-1}\!\left(\mathbf{x}_i^k, \mathbf{Z}_i(\mathbf{x}_i^k); \mathbf{K},
\mathbf{T}_i\right), \\
\mathbf{p}_{w,j}^k = \pi^{-1}\!\left(\mathbf{x}_j^k, \mathbf{Z}_j(\mathbf{x}_j^k); \mathbf{K}, \mathbf{T}_j\right),
\end{aligned}
\end{equation}
where \(\pi^{-1}(\cdot)\) denotes the unprojection operator.
We then minimize the confidence-weighted robust distance between matched 3D points:
\begin{align}
\mathcal{L}_{3D}
=
\sum_{i<j}\sum_{(\mathbf{x}_i^k,\mathbf{x}_j^k)\in \mathbf{M}_{i\rightarrow j}^v}
\mathbf{Q}^v_{i\rightarrow j}(\mathbf{x}_i^k,\mathbf{x}_j^k) \\
\rho_{\delta_{3D}}
\Big(
\|\mathbf{p}_{w,i}^k-\mathbf{p}_{w,j}^k\|_2
\Big),
\end{align}
where $\rho_{\delta_{3D}}(\cdot)$ is the Huber robust penalty with transition threshold $\delta_{3D}$.

\paragraph{Stage II: 2D reprojection consistency}
After the 3D alignment converges, we further tighten multi-view consistency by enforcing symmetric reprojection agreement. Specifically, we project \(\mathbf{p}_{w,i}^k\) into view \(j\) and \(\mathbf{p}_{w,j}^k\) into view \(i\):
\begin{equation}
\hat{\mathbf{x}}_{i\rightarrow j}^k = \pi\!\left(\mathbf{p}_{w,i}^k; \mathbf{K}, \mathbf{T}_j^{-1}\right), 
\hat{\mathbf{x}}_{j\rightarrow i}^k = \pi\!\left(\mathbf{p}_{w,j}^k; \mathbf{K}, \mathbf{T}_i^{-1}\right),
\end{equation}
and minimize the confidence-weighted robust reprojection error:
\begin{equation}
\begin{aligned}
\mathcal{L}_{2D}
=
\sum_{i<j}
\sum_{(\mathbf{x}_i^k,\mathbf{x}_j^k)\in \mathbf{M}_{i\rightarrow j}^v}
\mathbf{Q}^v_{i\rightarrow j}(\mathbf{x}_i^k,\mathbf{x}_j^k)\\
\Big[
\rho_{\delta_{2D}}\!\left(\|\hat{\mathbf{x}}_{i\rightarrow j}^k-\mathbf{x}_j^k\|_2\right)
+
\rho_{\delta_{2D}}\!\left(\|\hat{\mathbf{x}}_{j\rightarrow i}^k-\mathbf{x}_i^k\|_2\right)
\Big],
\end{aligned}
\end{equation}
where \(\rho_{\delta_{2D}}(\cdot)\) is the Huber loss with with transition threshold $\delta_{2D}$.

The overall refinement objective is optimized in an two-stage manner:
\begin{equation}
\left\{
\begin{aligned}
\text{Stage I:}\quad & \min_{\mathbf{Z}} \ \mathcal{L}_{3D}
\quad \text{for } N_{3D}\ \text{iters},\\
\text{Stage II:}\quad & \min_{\mathbf{Z}} \ \mathcal{L}_{2D}
\quad \text{for } N_{2D}\ \text{iters}.
\end{aligned}
\right.
\end{equation}
We denote $\mathbf{Z}^*$ as our final refined depth maps.

\begin{table*}[h]
\tiny
\renewcommand{\arraystretch}{1.4}
\tabcolsep=0.2cm
\caption{Results on GraspNet Dataset, showing AP on the RealSense/Kinect splits respectively}
\vspace{-0.4cm}
\begin{center}
\resizebox{0.85\textwidth}{!}{
\begin{tabular}{c|c|c|c|c|c} 
\hline
\textbf{Method} & \textbf{Data} & \textbf{Seen} & \textbf{Similar} & \textbf{Novel} & \textbf{Average} \\
\hline
PointnetGPD\cite{liang2019pointnetgpd}~ & RGB-D & 25.96/27.59 & 22.68/24.38 & 9.23/10.66 & 19.29/20.88 \\
GraspNet\cite{fang2020graspnet}~ & RGB-D & 27.59/29.88 & 26.11/27.84 & 10.55/11.51 & 21.41/23.08 \\
TransGrasp\cite{liu2022transgrasp}~ & RGB-D & 39.81/35.97 & 29.32/29.71 & 13.83/11.41 & 27.65/25.70 \\
HGGD\cite{chen2023efficient}~ & RGB-D & 59.36/60.26 & 51.20/48.59 & 22.17/18.43 & 44.24/42.43 \\
FlexLog\cite{xie2025rethinking}~ & RGB-D & \textbf{72.81}/\textbf{69.44} & \textbf{65.21}/\textbf{59.01} & \textbf{30.04}/\textbf{23.67} & \textbf{56.02}/\textbf{50.67} \\

\hline
GraspNeRF\cite{dai2023graspnerf}~ & RGB & 22.49/24.61 & 14.15/17.67 & 11.08/12.86 & 15.91/18.38 \\
VG-Grasp\cite{dai2026visual} & RGB & 59.23/54.65 & 36.34/35.13 & 10.84/11.85 & 35.47/33.88 \\
\textbf{Ours} & RGB & \textbf{63.70}/\textbf{66.80} & \textbf{56.03}/\textbf{57.35} & \textbf{23.22}/\textbf{20.47} & \textbf{47.65}/\textbf{48.21} \\

\hline
\end{tabular}
}  
\end{center}
\label{tab:grasp_results}
\vspace{-0.5cm}
\end{table*}

\textbf{Grasp-aware Guidance.}
After refinement, we fuse all views into a consistent colored point cloud, but we do not unproject all pixels. We define the valid matching Mask of image $i$, $\mathbf{M}^v_i \in \{0,1\}^{H\times W}$ as: 
\begin{equation}
\mathbf{M}^v_i(\mathbf{x})
=
\mathbf{1}\!\left(
\exists j\neq i,\ \exists (\mathbf{x},\mathbf{x}_j)\in \mathbf{M}^v_{i\rightarrow j}
\right).
\end{equation}
This mask is further intersected with the initial confidence mask $\hat{\mathbf{C}}_i$ and the MobileSAM\cite{zhang2023faster} segmentation mask $\mathbf{S}_i$ (for background points removal), yielding the final valid pixel $\mathbf{V}_i \in \{0,1\}^{H\times W}$ as:
\begin{equation}
\mathbf{V}_i(\mathbf{x})
=
\mathbf{M}^v_i(\mathbf{x})\odot 
\left(\hat{\mathbf{C}}_i(\mathbf{x})>\tau_C\right)\odot 
\mathbf{S}_i(\mathbf{x}),
\end{equation}
where $\tau_C$ denotes pointmap confidence threshold. We unproject only pixels with $\mathbf{V}(\mathbf{x})=1$ using the refined depth $\mathbf{Z}^*$, transform them to the world frame, and fuse different estimates to get multi-view consistent point cloud $\mathcal{P}$. Finally, we remove isolated outliers by a radius-based neighbor filter and get point cloud as:
\begin{equation}
\mathcal{P}_{\mathrm{filt}}
=
\left\{
\mathbf{p}\in\mathcal{P}\ \big|\
\left|\left\{\mathbf{q}\in\mathcal{P}:\|\mathbf{q}-\mathbf{p}\|_2<r\right\}\right|
\ge n_{\min}
\right\},
\end{equation}
where $r$ and $n_{\min}$ denote the radius threshold and the minimum number of neighbors, respectively.

\textbf{Grasp Generation.}
Given the fused point cloud $\mathcal{P}_{\mathrm{filt}}$ transformed to camera frame, we adopt the Local Grasp (LoG) module~\cite{xie2026rethinking} and perform grasp detection in a grasp-centric, region-level manner. We first generate $N_c$ region centers via Furthest Point Sampling (FPS) on $\mathcal{P}_{\mathrm{filt}}$, and then aggregate a local region around each center using a ball query. Conditioned on each local point cloud, LoG predicts multiple region-level grasp candidates
$\mathbf{g}_p=(\Delta \mathbf{t}, \boldsymbol{\theta}, w)$,
where $\Delta\mathbf{t}\in\mathbb{R}^3$ refines the grasp center relative to the sampled region center, $\boldsymbol{\theta}\in\mathbb{R}^3$ denotes the gripper orientation in Euler angles, and $w$ is the gripper width. We train the LoG module based on our filterd point cloud. Each candidate is converted to a 6-DoF grasp configuration in the world frame as
$\mathbf{g}=(\mathbf{t},\boldsymbol{\theta},w)$ with an associated quality score $s$.
Finally, we apply non-maximum suppression in $\mathrm{SE(3)}$ and retain the top-ranked grasps for execution.

\section{EXPERIMENTS}

We evaluate the effectiveness of our framework through comprehensive tests on both the GraspNet-1Billion\cite{fang2020graspnet} real dataset and physical robotic grasping experiments. Comparative experiments were conducted using heterogeneous data captured by Realsense and Kinect depth cameras. We only use the \textbf{RGB} information.

\subsection{Implementation Details}
We conduct experiments on GraspNet-1Billion and a real-world robotic setup. For each scene, we select $N{=}5$ RGB views using a coverage-oriented sampling strategy with mild randomness, and use the provided camera intrinsics $\mathbf{K}$ and poses $\mathbf{T}_i$. The matching correspondence threshold $\tau_Q$ is set to 3.0 and pointmap confidence threshold $\tau_C$ is set to 3.0. In depth refinement, Stage I iteration $N_{3D}$ is set to 100 and Stage II iterarion $N_{2D}$ is set to 100. The Huber transition thresholds $\delta_{3D}$ and $\delta_{2D}$ are set to 0.3, 3.0 respectively. For radius-based point cloud filtering, radius $r$ is set to 0.03 and minimum neighbor count $n_{\min}$ is set to 60. For the MobileSAM\cite{zhang2023faster} segmentation, we implement ViT-Tiny model. For grasp generation, the number of region centers $N_c$ is set to 500. We use the LoG module~\cite{xie2026rethinking}. In real-world experiments, we use predefined viewpoints (fixed poses) and $N{=}4$ views per scene. $\tau_Q$ is set to 1.0 and $\tau_C$ is set to 1.0, to accommodate real-image noise and appearance variation. $N_c$ is set to 100, $r$ is set to 0.05 and $n_{\min}$ is set to 300 to suppress sparse outliers.

The MG-Grasp system is implemented on a desktop computer with an NVIDIA RTX 4090 GPU and Intel Core i9-13900K CPU. Our system runs at seconds-level end-to-end latency of 2.86s per inference with 5 views and 2.13s per inference with 4 views.

\subsection{GraspNet Performance Evaluation}
\label{sec:performance}

We evaluate our method on the GraspNet-1Billion benchmark following the official protocol, reporting Average Precision (AP) on three standard splits: \textbf{Seen}, \textbf{Similar}, and \textbf{Novel}, under both RealSense and Kinect settings. We compare against representative RGB-D 6-Dof grasp methods as well as recent RGB-based pipelines. Notably, as the original GraspNeRF~\cite{dai2023graspnerf} lacked validation on the GraspNet benchmark, we reimplement this model and perform end-to-end training on the standard dataset to ensure fair comparison.

Table~\ref{tab:grasp_results} reports AP on GraspNet. Overall, our RGB-only method consistently outperforms prior RGB-based grasping approaches on average scores on RealSense/Kinect cameras. More specifically, compared with NeRF-based grasping (GraspNeRF~\cite{dai2023graspnerf}), our method achieves higher AP across all splits, with an average improvement of \textbf{+31.74/+29.83} AP on RealSense/Kinect. We attribute this gain to our more accurate geometry representation. Compared with VG-Grasp, which directly feeds point clouds generated by DUSt3R\cite{wang2024dust3r} into a downstream grasp module, our method further improves the average AP by \textbf{+12.18/+14.33}. This supports the effectiveness of our \textbf{Metric-Scale Depth Map Generator} and  \textbf{Multi-View Guided Grasp Generation} modules. Notably, our method narrows the gap to the strong RGB-D upper bound FlexLoG~\cite{xie2026rethinking}, despite using no depth information. 
\begin{table}[t]
\centering
\small
\setlength{\tabcolsep}{6pt}
\renewcommand{\arraystretch}{1.15}
\caption{Results of different geometry pipelines on GraspNet}
\label{tab:geomotry_diff}
\resizebox{0.48\textwidth}{!}{
\begin{tabular}{l c c c c}
\toprule
\textbf{Method} & \textbf{Seen} & \textbf{Similar} & \textbf{Novel} & \textbf{Average} \\
\midrule
\makecell[l]{FeedForward-based  \\ (VGGT\cite{wang2025vggt})} & 52.29/51.54 & 47.98/43.18 & 20.73/\textbf{22.90} & 40.33/39.21 \\
\midrule
\makecell[l]{SLAM-based \\
(MASt3R-SLAM\cite{murai2025mast3r})} & 50.39/43.42 & 38.54/37.50 & 18.19/14.10 & 35.71/31.67 \\
\midrule
\makecell[l]{MVS-based \\
(MVSAnywhere\cite{izquierdo2025mvsanywhere})} & 42.64/39.12 & 33.92/32.41 & 14.83/12.78 & 30.46/28.10 \\
\midrule
\textbf{Ours} & \textbf{63.70}/\textbf{66.80} & \textbf{56.03}/\textbf{57.35} & \textbf{23.22}/20.47 & \textbf{47.65}/\textbf{48.21} \\
\bottomrule
\end{tabular}}
\end{table}

\begin{figure}[h]
  \centering
  \includegraphics[width=0.5\textwidth]{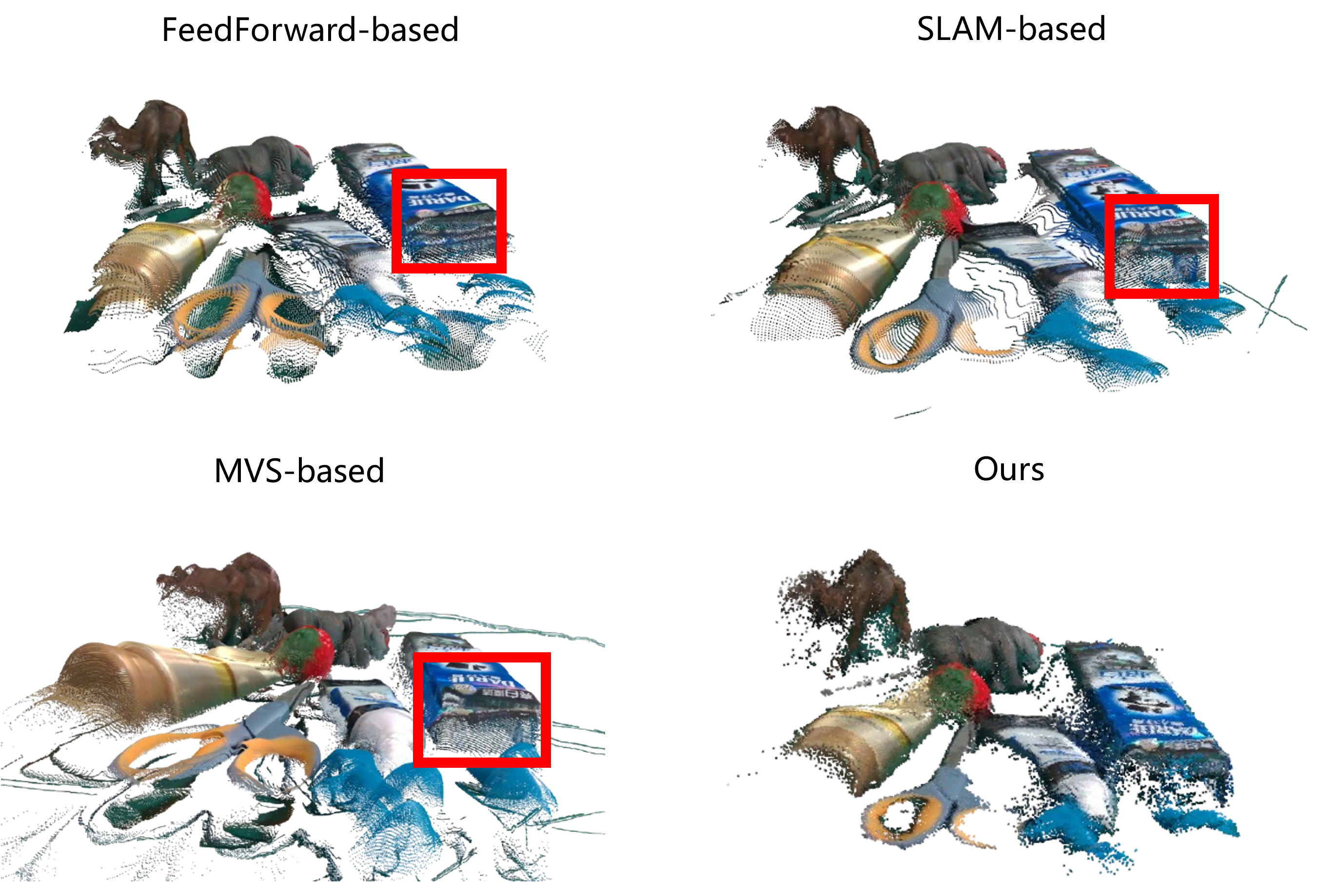}
  \caption{Comparison of unprojected point clouds produced by different geometry pipelines. The highlighted regions illustrate misaligned surface.}
  \label{backbone_diff}
  \vspace{-3mm}
\end{figure}

To verify that the gains of MG-Grasp are not simply due to the choice of a stronger geometry backbone, we construct three baselines by pairing the same LoG grasp decoder with three different geometry backbones: \textbf{FeedForward-based} predictor (VGGT~\cite{wang2025vggt}), \textbf{SLAM-based} system (MASt3R-SLAM~\cite{murai2025mast3r}), and \textbf{MVS-based} method (MVSAnywhere~\cite{izquierdo2025mvsanywhere}). For all baselines, LoG is kept identical and is fed with point clouds unprojected from the corresponding depth outputs. As shown in Table~\ref{tab:geomotry_diff}, we keep the LoG module identical and feed it with point clouds unprojected from the depths produced by different backbones. Since VGGT~\cite{wang2025vggt} and MASt3R-SLAM~\cite{murai2025mast3r} output up-to-scale geometry, we recover metric scale by aligning their estimated camera trajectory to the ground-truth poses using \texttt{evo}\cite{grupp2017evo} and rescale the predicted depths accordingly. As shown in Fig. \ref{backbone_diff}, although these methods can provide multi-view geometry, their reconstructed depths are often not sufficiently multi-view consistent under the given camera intrinsics/extrinsics, which directly degrades the region-level local grasp reasoning in LoG. In contrast, our method explicitly refines all depth maps with dense multi-view correspondences under metric-scale constraints, producing a grasp-oriented point cloud.

\subsection{Effect of Input View Count}
To study how visual input density affects reconstruction-driven grasping, we evaluate our framework using different numbers of input views, ranging from 2 to 9. For each GraspNet scene, we sample views with the coverage-oriented strategy. We report average AP under both RealSense and Kinect.

\begin{figure}[h]
  \centering
  \includegraphics[width=0.45\textwidth]{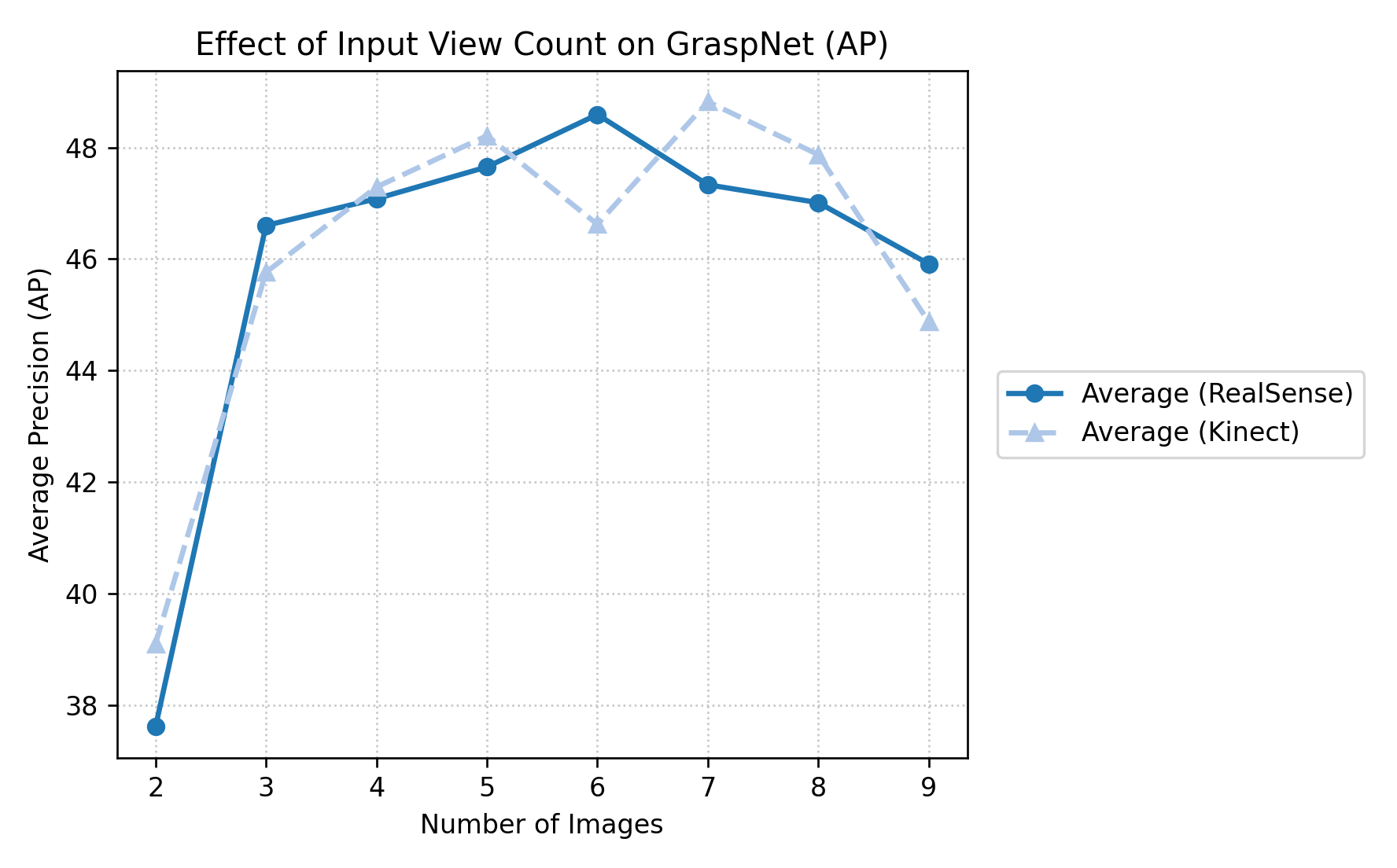}
  \caption{Effect of the number of input views on GraspNet grasping performance (AP).}
  \label{img_nums}
\end{figure}

As shown in Fig.~\ref{img_nums}, increasing the number of input images generally improves grasping performance, with the largest gains observed when moving from sparse inputs (2--3 views) to moderate multi-view inputs (4--6 views). In particular, using only two views leads to noticeably lower AP, indicating that limited viewpoints often yield incomplete geometry over the scene. Performance saturates around 5--6 views, suggesting that additional views beyond this range provide diminishing returns and may introduce redundant observations. Empirically, this range provides most of the performance gains of multi-view reconstruction while avoiding unnecessary pairwise constraints and runtime overhead from processing extra views.


\subsection{Ablation Study}
\label{sec:ablation}

We conduct ablation studies on GraspNet to validate the effectiveness of the two key components in MG-Grasp: (A) \textbf{Triangulation-based Scale Recovery}, and (B) \textbf{Multi-View Refinement} for multi-view consistency. 


\begin{table}[t]
\centering
\small
\setlength{\tabcolsep}{6pt}
\renewcommand{\arraystretch}{1.15}
\caption{Ablation study of different modules on GraspNet. 
(A) Triangulation-based scale recovery; (B) multi-view refinement. }
\label{tab:ablation}
\resizebox{0.48\textwidth}{!}{
\begin{tabular}{c c c c c c}
\toprule
(A) & (B) & \textbf{Seen} & \textbf{Similar} & \textbf{Novel} & \textbf{Average} \\
\midrule
\xmark & \xmark & 29.27/23.99 & 25.48/19.51 & 14.32/10.64 & 23.02/18.05 \\
\xmark & \cmark & 40.73/35.06 & 37.10/23.57 & 20.90/11.85 & 32.91/23.49 \\
\cmark & \xmark & 55.29/39.95 & 43.67/38.66 & 20.22/16.25 & 39.73/31.62 \\
\cmark & \cmark & \textbf{63.70}/\textbf{66.80} & \textbf{56.03}/\textbf{57.35} & \textbf{23.22}/\textbf{20.47} & \textbf{47.65}/\textbf{48.21} \\
\bottomrule
\end{tabular}}
\end{table}
Without scale recovery, the depth predictions remain up-to-scale and thus cannot be directly evaluated in metric grasping, leading to near-zero valid grasps. Therefore, for the variant without (A), we run Mast3r-sfm\cite{duisterhof2025mast3r} to estimate camera poses from the same RGB views and adopt the pose-alignment-based scale calibration using \texttt{evo}\cite{grupp2017evo}. As shown in Table \ref{tab:ablation}, Compared with this pose-alignment baseline, our triangulation-based scale recovery consistently yields better grasp AP across all splits. We further remove the proposed Multi-view refinement module and directly use the scaled initial depth maps for point cloud fusion and grasp generation. The results of variant without (B) show a clear performance drop, especially on \textbf{Seen} and \textbf{Similar} objects, suggesting that the refinement stage is crucial for enforcing cross-view geometric consistency and suppressing mismatched/outlier correspondences before point cloud fusion. This validates our design of using confidence-weighted multi-view optimization to bias geometric refinement toward robust grasp generation under sparse views.

\subsection{Real Robot Experiment}
\begin{figure}[h]
  \centering
  \includegraphics[width=0.3\textwidth]{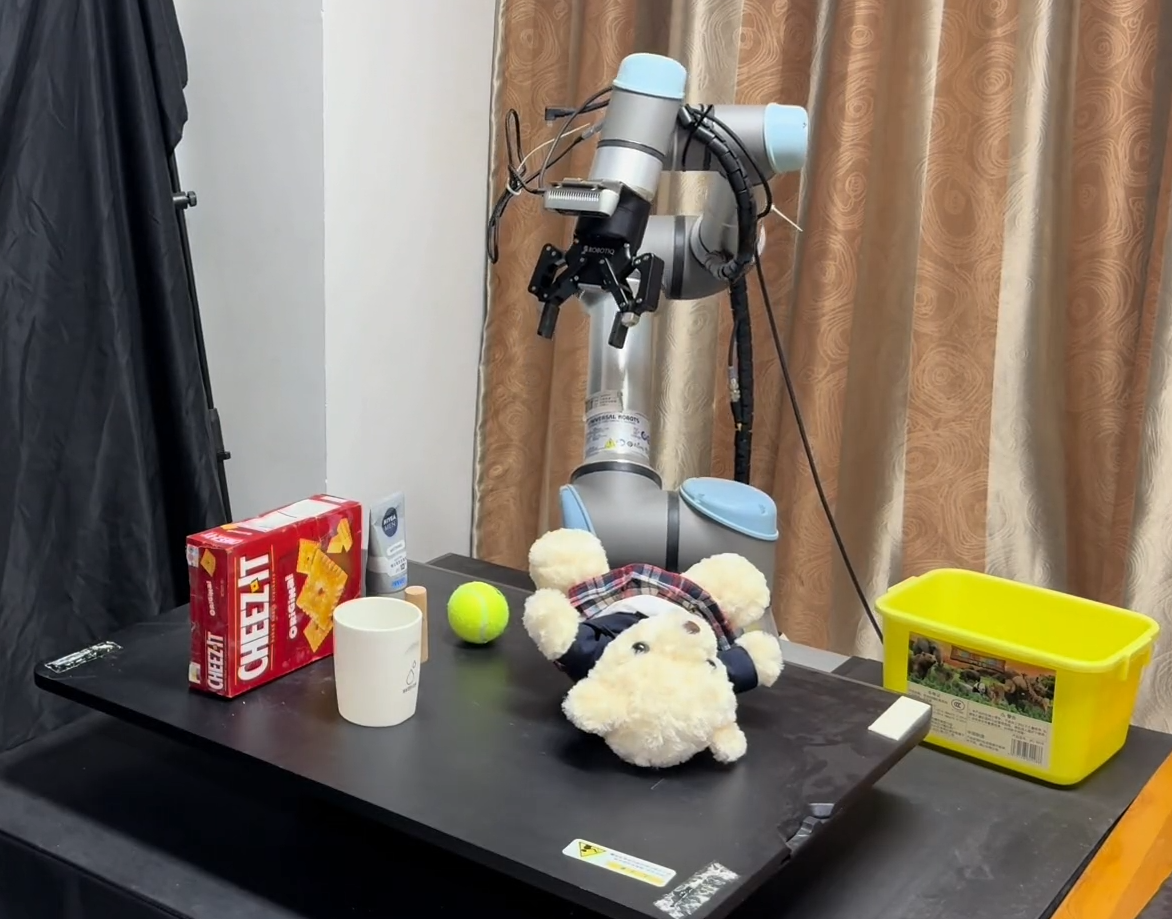}
  \caption{Real world setting.}
  \label{refinement_diff}
\end{figure}

We further validate MG-Grasp on a real robotic platform in tabletop scenes. The setup consists of a UR5e manipulator with a Robotiq 2F-85 adaptive parallel gripper and a RealSense D435i RGB-D camera. We only use the RGB information. Different from prior real-robot evaluations that typically rely on a single view, we acquire 4 sparse RGB views per scene to run our multi-view pipeline. All experiments are executed on a desktop equipped with an NVIDIA RTX 4090 GPU, and MG-Grasp runs with an end-to-end latency of about 2.13s.

\begin{table}[htbp]
\caption{REAL ROBOTIC GRASP EXPERIMENT RESULTS}
\label{tab:robotic_exp}
\centering
\begin{tabular}{ccc}
    \toprule
    \textbf{Scene}  & \textbf{Success} & \textbf{Attempt} \\
    \midrule
    1 & 5 & 6 \\
    2 & 5 & 5 \\
    3 & 5 & 5 \\
    4 & 5 & 6 \\
    5 & 5 & 6 \\
    6 & 5 & 6 \\
    7 & 5 & 6 \\
\midrule[0.8pt]
\multicolumn{1}{l}{Success Rate\textsuperscript{1}} & \multicolumn{2}{r}{35/40 = 87.5\%} \\
\multicolumn{1}{l}{Completion Rate\textsuperscript{2}} & \multicolumn{2}{r}{35/35 = 100\%} \\
\bottomrule
\end{tabular}

\vspace{0.5em}
\footnotesize
\textsuperscript{1} Success Rate = Total Success / Total Attempts \\
\textsuperscript{2} Completion Rate = Successful Trials / Total Trials
\end{table}

We evaluate MG-Grasp on a set of 25 daily-use objects with varied shapes and sizes. In each cluttered scene, five objects are randomly sampled and placed on a tabletop with arbitrary poses. For each trial, the robot iterates between grasp proposal and execution, terminating when (i) all objects that remain graspable are removed, or (ii) two consecutive grasp failures occur. Performance is measured by grasp success rate (successful lifts / total attempts) and task completion rate (cleared objects / total objects across all scenes).

Table~\ref{tab:robotic_exp} reports that MG-Grasp achieves $35/40=87.5\%$ grasp success and clears $35/35=100\%$ of objects. Despite the strong overall performance, the remaining failures mainly fall into two categories. First, highly reflective objects or near-spherical objects tend to produce unreliable dense correspondences, which increases matching errors and leads to inaccurate or incomplete reconstructed geometry. As a result, the downstream 6-DoF grasps can be offset in position or orientation. Second, for objects with unusual shapes, our randomly sampled grasp guidance may occasionally focus on less favorable regions, making it harder for the grasp generator to propose stable contacts.

Although MG-Grasp is not specifically designed for transparent objects, it still demonstrates promising performance in this setting, achieving an 11/20 = 55.0\% grasp success rate and an 11/15 = 73.3\% completion rate. Since depth sensors typically fail to produce reliable point clouds on transparent surfaces, these results suggest that our RGB-only pipeline has strong potential for efficient transparent-object grasping where RGB-D methods are inherently limited.

\subsection{Runtime Analysis}
\label{sec:runtime}

We report the average  runtime of each module in our pipeline, measured over 500 inference runs on the GraspNet RealSense/Kinect splits. As shown in Table~\ref{tab:runtime_breakdown}, the proposed system runs with a seconds-level latency of 2.86s per inference on average. The runtime is dominated by the geometry stage, where MASt3R\cite{murai2025mast3r} matching and the two-stage refinement together account for approximately 77.3\% of the total latency. This is mainly because we impose pairwise geometric constraints across image pairs, requiring dense correspondence extraction and iterative optimization over all coupled views, which increases the computational cost. These results suggest that further acceleration can be achieved by reducing the number of processed view pairs and iterations in the refinement stage.

We further compare the end-to-end runtime of MG-Grasp with representative RGB-only baselines under the same evaluation protocol. As reported in Table~\ref{tab:runtime_compare}, MG-Grasp is both more accurate and more efficient than VG-Grasp\cite{dai2026visual}, indicating a better accuracy and efficiency for RGB-only 6-DoF grasping. We also compare our runtime with FeedForward-based pipeline implemented in Sec. \ref{sec:performance}. Although FeedForward-based pipeline is significantly faster due to its single-pass prediction, it does not explicitly enforce consistency under known camera parameters, and thus yields less reliable geometry for downstream grasp reasoning, resulting in lower grasping performance than MG-Grasp.

\begin{table}[t]
\centering
\small
\setlength{\tabcolsep}{6pt}
\renewcommand{\arraystretch}{1.15}
\caption{Runtime breakdown of MVR-Grasp per inference}
\label{tab:runtime_breakdown}
\resizebox{0.48\textwidth}{!}{
\begin{tabular}{lcc}
\toprule
\textbf{Module} & \textbf{Time (s)} & \textbf{Ratio (\%)} \\
\midrule
\textbf{MobileSAM masking} & 0.06 & 2.1 \\
\textbf{Depth Aggregation and Correspondence} & 0.87 & 30.5 \\
\textbf{Triangulation-based Scale Recovery} & 0.11 & 3.8 \\
\textbf{Multi-View Refinement} & 1.34 & 46.8 \\
\textbf{Grasp-aware Guidance} & 0.05 & 1.6 \\
\textbf{Grasp Generation} & 0.30 & 10.6 \\
\midrule
\textbf{Total} & \textbf{2.86} & \textbf{100.0} \\
\bottomrule
\end{tabular}}
\end{table}
\begin{table}[t]
\centering
\small
\setlength{\tabcolsep}{7pt}
\renewcommand{\arraystretch}{1.0}
\caption{End-to-end runtime comparison (seconds per inference).}
\label{tab:runtime_compare}
\begin{tabular}{lc}
\toprule
\textbf{Method} & \textbf{Time (s)} \\
\midrule
VG-Grasp\cite{dai2026visual} & 11.20 \\
\midrule
\makecell[l]{FeedForward-based  \\ 
(VGGT\cite{wang2025vggt})} & 0.61 \\
\midrule
\textbf{Ours} & 2.86 \\
\bottomrule
\end{tabular}
\end{table}
Finally, we analyze the runtime in the real-robot setting. One complete grasp cycle, including viewpoint motion, grasp execution and object placement, takes 32.57s on average. Within this cycle, MG-Grasp contributes 2.13s for end-to-end 6-DoF grasps inference, accounting for only a small fraction ($\approx$6.5\%) of the total task time. This indicates that, despite the seconds-level inference latency, the overall system throughput in our real-robot experiments is mainly constrained by robot motion and manipulation execution rather than perception alone.

\section{CONCLUSIONS}

We present MG-Grasp, a depth-free 6-DoF grasping framework that derives high-quality grasps directly from sparse multi-view RGB observations. Instead of treating reconstruction as a separate preprocessing step, MG-Grasp couples metric scale recovery, multi-view consistency refinement, and grasp generation in a unified pipeline, leading to stronger performance than prior RGB-only baselines. Notably, MG-Grasp achieves performance comparable to representative RGB-D grasping pipelines, while operating without depth sensors. Extensive evaluations further demonstrate practical effectiveness in tabletop scenes. Future work will focus on accelerating the pipeline through more efficient multi-view selection, lighter refinement, and GPU-friendly fusion to enable faster closed-loop grasping.

\bibliographystyle{IEEEtran}
\bibliography{reference}
\end{document}